\documentclass{article}
\usepackage{stywhispers,amsmath,epsfig}
\usepackage{pdflscape}
\usepackage{subfig}
\usepackage{amssymb}
\usepackage{graphicx}  
\usepackage{listings}
\usepackage{wrapfig}
\usepackage{enumerate}
\usepackage{epstopdf}
\usepackage{array}
\usepackage{enumerate}
\usepackage{multirow}
\usepackage{hhline}
\usepackage{color}
\usepackage{url}
\usepackage{color}

\newcommand{\norm}[1]{\lVert #1 \rVert}

\title{Classification of Hyperspectral Imagery on Embedded Grassmannians} 
\name{Sofya Chepushtanova and Michael Kirby
\thanks{This paper is based on research partially supported by the National Science Foundation
under Grants No. DMS-1228308 and DMS-1322508 as well as the DOD-USAF under Award Number FA9550-12-1-0408. 
Any opinions, findings, and conclusions or recommendations expressed in this material are 
those of the authors and do not necessarily reflect the views of the National Science Foundation 
or the United States Air Force.}
}
\address{Department of Mathematics, Colorado State University, Fort Collins, Colorado USA}
\begin{document}
\ninept
\maketitle
\begin{abstract} 
We propose an approach for capturing the signal variability in hyperspectral imagery
using the framework of the Grassmann manifold.  Labeled points from each class are sampled and
used to form abstract points on the Grassmannian.
The resulting points on the Grassmannian have representations as 
orthonormal matrices and as such do not reside in Euclidean space
in the usual sense. There are a variety of metrics which allow
us to determine a distance matrices that can be used
to realize the Grassmannian as an embedding in Euclidean space.  
We illustrate that we can achieve an approximately isometric embedding
of the Grassmann manifold using the chordal metric 
while this is not the case with geodesic distances.  
However, non-isometric embeddings
generated by using a pseudometric on the Grassmannian lead 
to the best classification results.
We observe that as the dimension of the Grassmannian grows, the accuracy of the classification 
grows to 100\% on two illustrative examples.  We also observe a decrease
in classification rates if the dimension of the points on the Grassmannian is
too large for the dimension of the Euclidean space. 
We use sparse support vector machines to perform additional model reduction. 
The resulting classifier selects a subset of dimensions of the embedding without loss in 
classification performance.
\end{abstract}

\section{Introduction}
\label{intro}
Hyperspectral imagery can be viewed from a data perspective as three-way arrays, 
often referred to as data cubes.  
The spectral bands that make up these cubes can be viewed as
high-dimensional color spaces extending the three colors 
of visual photography to $n$ dimensions where
$n$ can vary from tens to hundreds depending on the acquisition hardware.  
A set of $k$ pixels from the same data class 
can be characterized as a $k$-dimensional subspace of ${\mathbb R}^n$.
The key feature of the set of $k$ pixels is that they capture
variation in the class signal that is typical of its representation. 
The central idea of this paper is a novel algorithmic
framework for  capturing the geometry of the Grassmannian
in Eulcidean space for the purposes of representing the
variability of information in hyperspectral imagery.

We propose to encode observations of interest as subspaces.
A collection of subspaces has a natural mathematical
structure known as the Grassmann manifold.   
In mathematics this is referred to as an {\it abstract manifold} since it does not
reside in Euclidean space, i.e., it's properties are not 
described by $n$-tuples whose distances are measured via inner products.  
Recently it has become an active area of research to develop computational algorithms on non-Euclidean
spaces \cite{absil,edelman}.

Nash's famous {\it isometric embedding theorem} shows under what conditions Riemannian manifolds, 
Grassmann manifolds are a special case of these,  
can be embedded into Euclidean space such that the distances between points on the
manifold are preserved \cite{nash}.  
There are some interesting counter examples to this, e.g., the polar cap of a sphere
can not be embedded isometrically into the plane, distances between point must be distorted.

One approach for embedding subspaces into Euclidean space is 
multidimensional scaling (MDS), see, e.g.,  \cite{mardia} and references therein.
The result is a configuration of points in Euclidean 
space whose distances approximate the distances measured
on the abstract manifold.  
One can also establish in certain circumstances that the distances are preserved,
i.e., the embedding is isometric.

Geometric approaches have been proposed for
characterizing data on manifolds, i.e., 
nonlinear objects that behave locally like Euclidean space.  These 
data driven approaches for manifold
learning include, e.g.,  isometric mapping (ISOMAP) \cite{ISOMAP},
local linear embedding (LLE) \cite{LLE}, and Laplacian Eigenmaps \cite{belkin2003laplacian}.  
A number of practical algorithms based on ISOMAP and LLE have been
proposed for  applications to hyperspectral imagery, see, e.g., \cite{bach}. 
We note that in these methods, a manifold coordinate system is derived 
from computing the geodesic distances between the hyperspectral pixels, 
i.e., they are algorithms operating in pixel space. 
The algorithms applied to pixel space are using 
manifolds as a model for the data.  
In our approach, we first encode sets of pixels as subspaces 
which are viewed as points on a Grassmann manifold,
the existence of which is theoretically guaranteed. 
The Grassmann manifold is then embedded into Euclidean space using MDS.  
As we will describe below, this setting then permits the simultaneous classification and 
dimension selection performed in the Euclidean space.

The outline of this paper is as follows. 
In Section~\ref{frame} we describe the mathematical framework
behind encoding collections of pixels as subspaces using the geometry of the Grassmann manifold. 
In Section~\ref{MDS} we outline the methodology for approximating an isometric embedding of the Grassmannian.
The algorithm is summarized in Section~\ref{algorithm} and the computational results discussed in Section~\ref{comp}.
We summarize our findings in Section~\ref{conc}.

\section{Framework}
\label{frame}
The real Grassmann manifold (Grassmannian) $G(k,n)$ is the collection of all $k$-dimensional subspaces 
of $\mathbb{R}^{n}$, for fixed $k \leq n$.  
It has dimension $k(n-k)$ where in our case $n$ is the number of wavelengths (spectral bands)
and $k$ is the number of pixels (taken from the same class) 
to be used to define the subspaces.  
Points on $G(k,n)$ are represented by orthonormal matrices.  
These points are actually equivalence classes in the sense that any 
two bases which have the same span are viewed as the same point on the Grassmannian.
The Grassmannian characterizes $k$-dimensional subspaces
of $n$-dimensional Euclidean space as a quotient space of the orthogonal group, 
$G(k,n) = \frac{O(n)}{O(k) \times O(n-k)}.$
Note that this is in contrast to other methods for 
representing subspaces, e.g., Stiefel manifolds \cite{edelman}.

The power of the Grassmann manifold lies in its ability to provide a geometric framework for characterizing the relationship between subpaces.
In particular, there is a rich geometric structure that permits a variety of metrics
to be employed.  We briefly summarize the basic facts here; for additional details see \cite{chang}.

The Grassmann manifold is endowed with a Riemannian structure which affords the computation of distances 
between the points on the manifold using, e.g.,
geodesics (i.e. curves of the shortest length). 
There are several different definitions of the distance on $G(k,n)$ used in the literature.
The geodesic and chordal distances \cite{conway} between two subspaces $\mathcal{P}_1$ and $\mathcal{P}_2$ 
are given by $$d_g(\mathcal{P}_1,\mathcal{P}_2) = (\sum_{i=1}^k \theta_i^2)^{1/2} = \|{\theta}\|_2$$ and 
$$d_c(\mathcal{P}_1,\mathcal{P}_2) = (\sum_{i=1}^k (\sin \theta_i)^{2})^{1/2} = \|\sin{\theta}\|_2,$$ respectively, 
where $\theta = [\theta_1, \theta_2, \ldots, \theta_k]$
($0 \leq \theta_1 \leq \theta_2 \leq \ldots \leq \theta_k \leq \pi /2$)
is the vector of the principal vectors between the subspaces $\mathcal{P}_1$ and $\mathcal{P}_2$.
Note that principal angles are computed using SVD-based algorithm 
applied to the matrices $P_1$ and $P_2$ that are orthonormal bases for the subspaces $\mathcal{P}_1$ and $\mathcal{P}_2$, respectively \cite{bjorck}. 
We also define a pseudometric given by $d_p = \theta_1$ (the smallest principal angle) to be used in our computations; see also \cite{chang}.

There are several options for determining an abstract point
of the Grassmannian from data assembled as columns of a matrix
$Y\in\mathbb{R}^{n \times k}$.  
The most inexpensive is the Gram-Schmidt algorithm for constructing a QR decomposition.
The {\it thin} Singular Value Decomposition (SVD) 
of $Y = U \Sigma V^T$ provides a characterization of data
in terms of the $n \times k$ orthonormal matrix $U$.  
Given the column space of $Y$, $\mathcal{R}(Y)$, is spanned by the columns of $U$,  
for computational purposes we can choose $U$ to be a point on the Grassmannian.  

Suppose we constructed $p$ points on $G(k,n)$, $U_1$, $U_2$, \ldots, $U_p$, 
each associated with a set of pixels from the same class, using the method described above.
We can now obtain a symmetric distance matrix $D \in \mathbb{R}^{p \times p}$, 
with $D_{ii} = 0$ and $D_{ij} \geq 0$ being equal to one of the distances between $U_i$ and $U_j$.


\section{Embedding the Grassmann Manifold into Euclidean Space}
\label{MDS}
In this section we describe how to realize a configuration of a collection of $k$-dimensional 
subspaces as an embedding $d$-tuples where the appropriate size of the Euclidean space $d$
is to be determined from the data.

We assume that we have obtained a collection of abstract points on the Grassmannian $G(k,n)$.
These points are generated by subspaces of the sampled data.  
In practice, $n$ is fixed to be the number of spectral bands
and $k$ can vary as we change the size of our collection of similar points. 
Although one could consider subspaces of different dimension, we assume in this 
paper that all points are on the same Grassmannian.

Given a collection of points on $G(k,n)$, the next step is to produce a configuration of points
in Euclidean space (of some dimension to be determined)
that faithfully captures the interpoint distances, i.e., an isometric embedding.  
An attractive tool for accomplishing this is multidimensional scaling (MDS).

MDS is able to produce a configuration of points in a Euclidean space that captures the (possibly non-Euclidean)
distances  of the original data in an optimal manner.
It is worth emphasizing the we are measuring distances in the embedded Euclidean space as usual, 
but that they are approximating distances that were measured on a manifold.
Note again, that this is not always possible -- spheres can't be isometrically embedded in planes. 
Given a distance matrix $D \in \mathbb{R}^{p \times p}$, the basic steps of MDS are:

\begin{enumerate}
   \item Compute $B = H A H$, where $H = I-\frac{1}{p}ee^T$ and $A_{ij} = -\frac{1}{2} D_{ij}^2$ ($e$ is a vector of $p$ ones).
   \item Compute the spectral decomposition $B = \Gamma \Lambda \Gamma^T$. 
   \item Set $X := \Gamma \Lambda ^ {\frac{1}{2}}$. 
  \end{enumerate}  

$X$ is a configuration of points in $\mathbb{R}^d$, 
where $d=rank(B)=rank(X) \leq p-1$.
If $B$ is positive semi-definite, 
then there exists a configuration of points in the Euclidean space associated with the positive eigenvalues
whose interpoint distances are given by $D$ \cite{mardia}.
In practice, in the case when $B$ is not positive 
semi-definite (i.e., negative eigenvalues occur), 
the Euclidean space associated with the positive
eigenvalues provides the optimal approximation 
of the configuration of points. 
      
\section{Algorithm}
\label{algorithm}

To obtain a collection of abstract points on the Grassmannian $G(k,n)$, 
we generate subspaces for a chosen dimension $k$; in practice we look at the effect
of the choice of $k$ on the classification results. 
We begin from sampling with replacement subsets of the data points, 
forming $n \times k$ 'tall-skinny' matrices $Y_i$. 
Note that each $Y_i$ contains only points from the same class. 
We also retain the appropriate classification labels 
for these matrices as they will be needed in classification step.
Then, as described in Section \ref{frame}, we select $U_i$ to be an orthonormal matrix that spans $\mathcal{R}(Y_i)$. 

Once the points on the Grassmannian have been embedded in a Euclidean space of appropriate dimension $d$ (Section~\ref{MDS}),
we employ a linear sparse support vector machine (SSVM) \cite{mang,chep} to perform classification in the Euclidean space.
An SSVM, or $\ell_1$-norm\footnote{For $\mathbf{x} \in {\mathbb R}^n$, $\norm{\mathbf{x}}_1 = \sum_{i=1}^n |x_i|$} SVM, 
has an attractive feature of being able to reduce the number of variables, 
i.e., the dimension of the data, as determined by the non-zero weights in the $\ell_1$-norm optimization problem.
In our case, the SSVM reduces the dimension of the embedding: 
non-zero weights in the decision function indicate optimal dimensions of the embedding out of $d$ dimensions
determined by the number of positive eigenvalues of $B$.
These optimal dimensions can be used for improving classification rates and embedding visualization. 
In this paper, we use a pre-defined tolerance to determine non-zero SSVM weights 
that indicate the optimal dimensions of the embedding. 

As noted in Section~\ref{intro}, our approach differs from spectral methods such as ISOMAP or LLE,
that derive a manifold coordinate system for original data sets.
In our approach, the data points are encoded as subspaces on a Grassmann manifold, 
and then embedded into Euclidean space optimally preserving distances between them.
We perform the basic steps in the following order:     
original data in ${\mathbb R}^n$ $\rightarrow$ points on $G(k,n)$  $\rightarrow$ embedding in ${\mathbb R}^d$  $\rightarrow$
SSVM classification and dimension selection. 

The algorithm below summarizes the proposed approach with more details: 

\medskip

\noindent
\underline{\textbf{Algorithm: Classification on Grassmannians}} 

\medskip
\noindent
 \textbf{Input:} A hyperspectral data matrix $\mathbf{X} \in \mathbb{R}^{n \times m} $, whose columns are pixels with
class labels $d_{i}$,  $i=1,\dots, m$,
subspace dimension $k$, number of points $p$ to be sampled on $G(k,n)$. 
 \begin{enumerate}
   \item Compute $p$ points $U_i$ on $G(k,n)$ by finding
orthonormal matrices that span $\mathcal{R}(Y_i)$. 
   \item Compute the pairwise distances between points on $G(k,n)$.
   \item Implement the MDS algorithm to obtain a configuration of points
in dimension $d$, where $d$ is the number of positive eigenvalues. 
   \item Train an SSVM to obtain classification accuracy and optimal dimensions in the Euclidean space.
 \end{enumerate}  
\textbf{Output:} Embedding of $G(k,n)$ into $\mathbb{R}^d$ and an SSVM model in the Euclidean space; 
subset of SSVM selected optimal dimensions.

\section{Computational Results}
\label{comp}

We illustrate the algorithm and the mathematical framework  using the well-known
hyperspectral AVIRIS Indian Pines data set\footnote{This
dataset can be downloaded from  
\url{http:/dynamo.ecn.purdue.edu/~biehl/MultiSpec.}}.
This data was collected by the Aiborne Visible/Infrared Imaging Spectrometer (AVIRIS) 
over a small agricultural area in Northwestern Indiana in 1992.
It consists of $145 \times 145$ pixels by $220$ bands from $0.4$ to $2.4 \mu m$. 
Due to availability of the ground truth, $10366$ pixels were prelabeled to be in one of the $16$ classes. 
The unlabeled background pixels were not used in our experiments. 
To limit the scope of our paper, we consider two two-class 
problems and one three-class classification problem.  
We preprocessed the data by mean centering, i.e., 
the mean of the data was subtracted from each pixel in the scene.
The data was randomly partitioned into $50 \%$ for training and $50 \%$ for testing.
For this data, the original input space dimension is $n = 220$.

In a typical 2-class experiment, to derive an embedding for a chosen dimension $k$, 
we assemble $100$ $n \times k$ matrices $Y_i$ from each class, 
which results in constructing $p=200$ points $U_i$ on $G(k,n)$. 
This set of points contains equal numbers of training and testing subspaces.  
Following the steps of the Algorithm, we obtain a distance matrix $D \in \mathbb{R}^{p \times p}$,
and then apply MDS to embed the points,
keeping track of the class labels of the subspaces.

\begin{figure}[!t]
  \centering
   \includegraphics[width=1\linewidth,height=6cm]{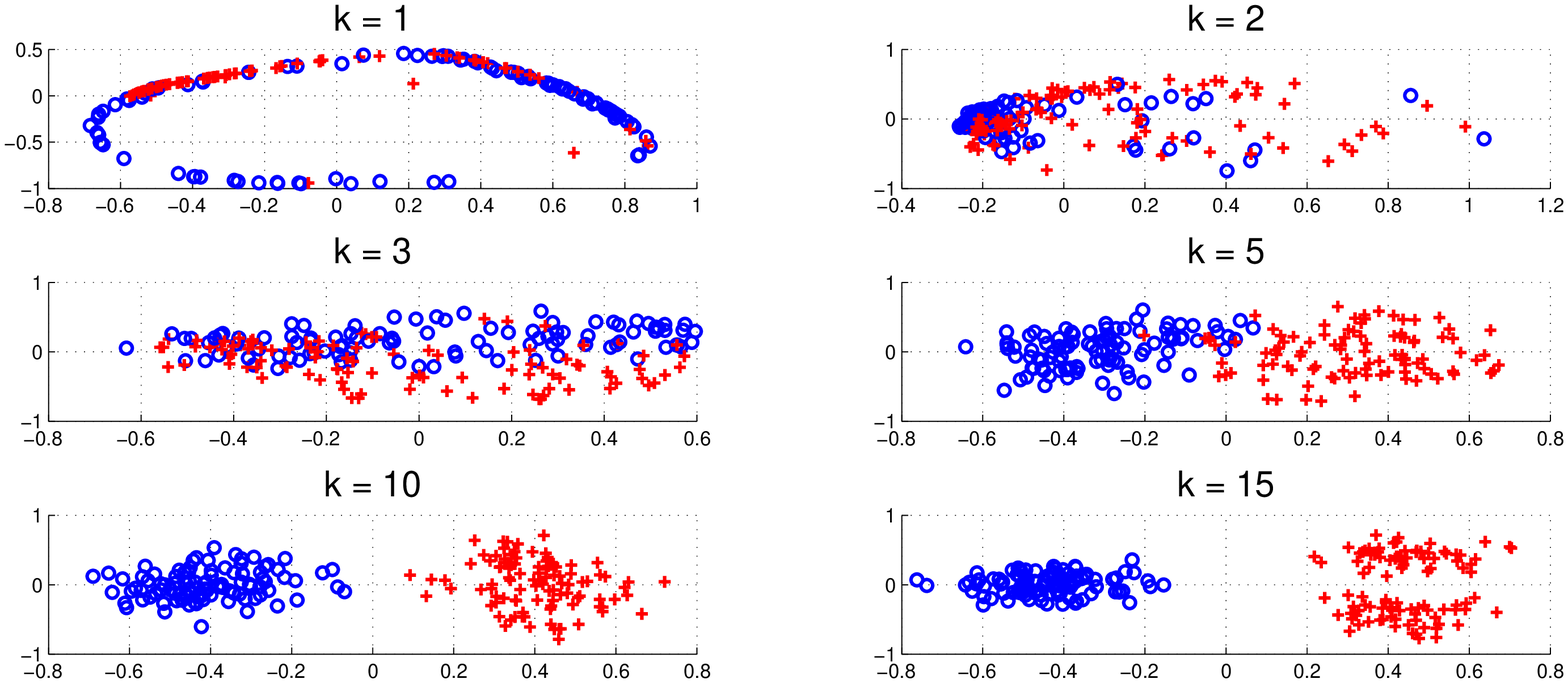}   
    \caption{Configuration of points on $G(k,220)$ embedded in Euclidean space using $d_g$ for 
     2 classes: Corn-notill (circles) and Grass/Pasture (plus signs). 
     Dimensions correspond to the two top eigenvalues of $B$ (MDS).}
   \label{figure_ExampleEmbedding1}
\end{figure}

\begin{figure}[!t]
  \centering
   \includegraphics[width=1\linewidth,height=6cm]{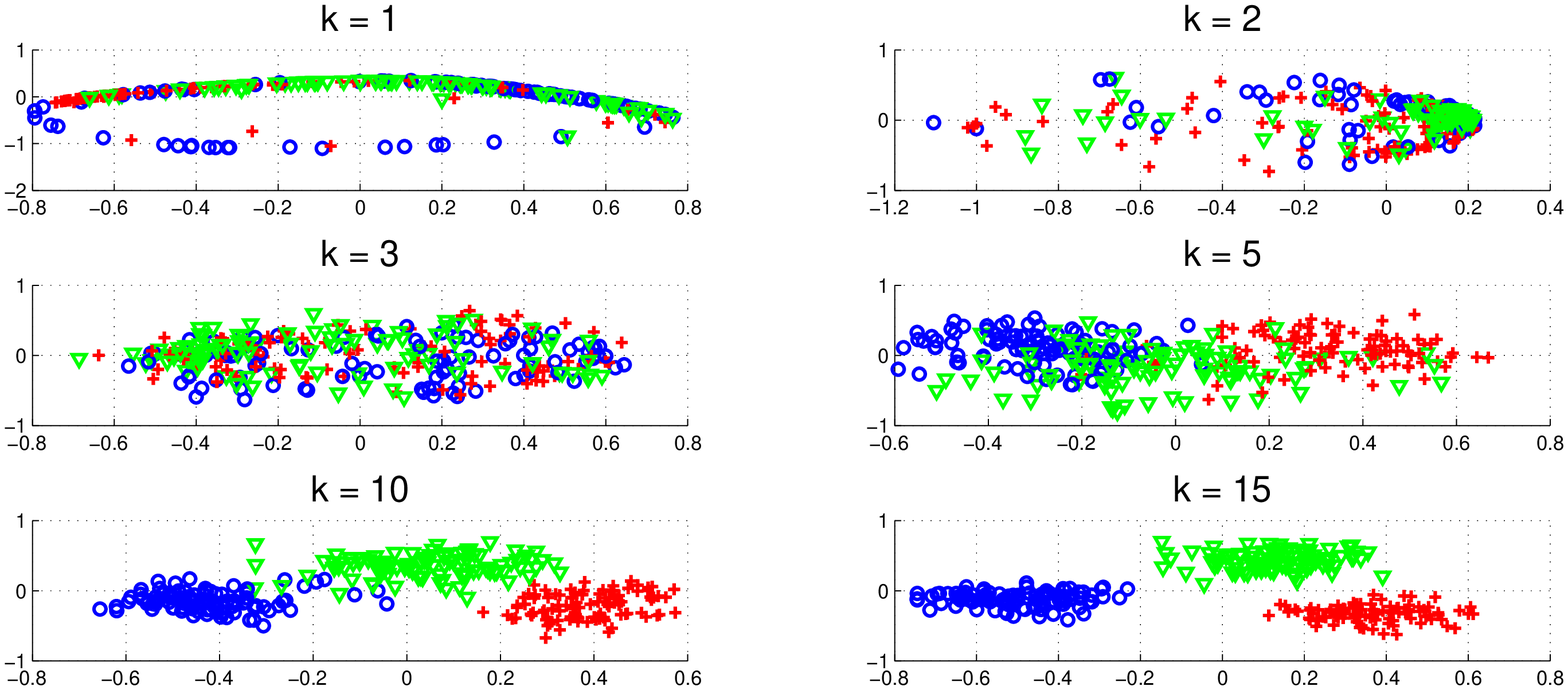}
    \caption{Configuration of points on $G(k,220)$ embedded in Euclidean space using $d_g$ for 
    3 classes: Corn-notill (circles), Grass/Pasture (plus signs), and Grass/Trees (triangles). 
    Dimensions correspond to the two top eigenvalues of $B$ (MDS).}
   \label{figure_ExampleEmbedding2}
\end{figure}

Figures \ref{figure_ExampleEmbedding1} and \ref{figure_ExampleEmbedding2} illustrate Euclidean embedding configurations
obtained for various $k$ using the geodesic distance $d_g$
for the two classes Corn-notill and Grass/Pasture and three classes Corn-notill, Grass/Pasture, and Grass/Trees, respectively.
Note that for the visualization purpose, 
the chosen two dimensions correspond to the two top eigenvalues of the matrix $B$ (MDS). 
We see the class separation becoming stronger as we
increase the dimension $k$ of the Grassmannians.
 
Table \ref{table_GrassmannResults1} contains accuracy rates and dimension selection results 
in the embedded spaces for the two classes Corn-notill and Grass-Pasture,
and Table \ref{table_GrassmannResults2} contains results for classes Soybeans-notill and Soybeans-min.
The former pair of classes has low classification difficulty in comparison to the latter.
The Soybean-notill and Soybeans-min classes were selected, 
because, to our knowledge, no-one has ever achieved $100\%$ classification accuracy on them.
Note that the results are averaged over 10 runs.  
For each $k$, we obtain a configuration of points in a new embedded space of dimension $d$, 
which corresponds to the number of positive eigenvalues of the matrix $B$.
We compare results obtained on embedded Grassmannians using two distance metrics, 
chordal $d_c$ and geodesic $d_g$, and the pseudometric $d_p$ defined as the smallest principal angle between subspaces.
Note that in case of the classes Corn-notill and Grass-Pasture, 
they are separated starting $k=10$ for all three choices of a metric. 
The best accuracy result we could obtain on the original data using SSVM was $99.27 \%$, 
while the Grassmannian framework allows one to improve the result up to $100 \%$. 
The number of dimensions selected by SSVM is very small for both chordal and pseudometric. 
As for the higher difficulty classification problem, namely, Soybeans-notill and Soybeans-min separation,
we also have improved the accuracy results using our framework. 
SSVM, applied to the original data, has $89.57 \%$, 
which is less than some accuracy results obtained using different metrics on the embedded Grassmannians. 
We observe that the performance of the pseudometric degrades from perfect 
to random as the number of dimensions is increase beyond the optimal number. 
This is a consequence of the fact that there are more dimensions in which to find a smallest single angle.

\begin{table*} [!ht]
\renewcommand{\arraystretch}{1.3}
\caption{Accuracy results and feature selection in embedded spaces for $p=200$ points on $G(k,220)$, Corn-notill versus Grass-Pasture, averaged over 10 runs:}
\label{table_GrassmannResults1}
\centering
\scalebox{0.82}{\centering
    \begin{tabular}{|c | p{1.5cm}| p{1.5cm}| p{1.5cm}| p{1.5cm}| p{1.5cm}| p{1.5cm}| p{1.5cm}| p{1.5cm}| p{1.5cm} |} 
          \hline   
        \bfseries Dimension of  subspaces $k$  & \multicolumn{3}{c|}{\bfseries Number of negative eigenvalues of $B$} & 
                                                \multicolumn{3}{c|}{\bfseries SSVM Accuracy ($\%$)} & 
                                         \multicolumn{3}{c|}{\bfseries Number of dimensions selected} \\ \hhline{~---------}        
       &  Chordal & Geodesic  & Pseudo  &  Chordal & Geodesic  & Pseudo &  Chordal & Geodesic & Pseudo \\ \hline     
    1  &    13.5    &   68.1   &   69.5 &   80.6   &   85.4    &  85.3  &   24.4   &   41.5   &  36.7 \\ 
    5  &     0      &   52     &   94.3 &   96.6   &   96      &  99.4  &    146   &  123.4   &  23.7 \\ 
    10 &     0      &   5.6    &   96.6 &   99.7   &   100     &  100   &    150.4 &  143.5   &  15.3  \\     
    15 &     0      &   0      &   98.2 &   100    &   99.8    &  100   &    160.6 &  172     &  7.6   \\
    20 &     0      &   0      &   99.6 &   100    &   100     &  100   &    161.1 &  164.9   &  5.2   \\
    25 &     0      &   0      &   99.7 &   100    &   98.2    &  100   &    162.7 &  171.9   &  5.4   \\ \hline  
   \end{tabular}
}
\end{table*}

\begin{table*}[!ht]
\renewcommand{\arraystretch}{1.3}
\caption{Accuracy results and feature selection in embedded spaces for $p=200$ points on $G(k,220)$, Soybeans-notill and Soybeans-min, averaged over 10 runs:}
\label{table_GrassmannResults2}
\centering
\scalebox{0.82}{\centering
    \begin{tabular}{|c | p{1.5cm}| p{1.5cm}| p{1.5cm}| p{1.5cm}| p{1.5cm}| p{1.5cm}| p{1.5cm}| p{1.5cm}| p{1.5cm} |} 
        \hline
        \bfseries Dimension of  subspaces $k$ & \multicolumn{3}{c|}{\bfseries Number of negative eigenvalues of $B$} & 
                                                \multicolumn{3}{c|}{\bfseries SSVM Accuracy ($\%$)} & 
                                         \multicolumn{3}{c|}{\bfseries Number of dimensions selected} \\ \hhline{~---------}    
       &  Chordal & Geodesic  & Pseudo  &  Chordal & Geodesic  & Pseudo &  Chordal & Geodesic & Pseudo \\ \hline     
    1  &   8.2    &   62.4    &  63.3   &    68    &    68.1  &   66.7  &    41.6  &   40.1   &  49    \\ 
    5  &   0      &   50.6    &  90.2   &    67.6  &    62.6  &   90.6  &   175.6  &  138.9   &  31.7  \\
    10 &   0      &   3.9     &  92.9   &    86.7  &    74.3  &   99.9  &    181.7 &   176.8  &  18.4  \\  
    15 &   0      &   0       &  94.5   &    94.2  &    86.6  &  100    &    175.8 &   179.3  &  10.5  \\ 
    20 &   0      &   0       &  96.2   &    99    &     92   &   95    &    169.5 &   181.2  &  4.9   \\
    25 &   0      &   0       &  97.3   &    93.7  &    92.8  &   50    &    166.3 &   172    &  3.3   \\ \hline  
   \end{tabular}
}
\end{table*}

\section{Conclusions}
\label{conc}
We propose an algorithm for analyzing hyperspectral imagery
using the geometric framework of the Grassmann manifold.  
Sets of same class pixels are encoded as points
on Grassmann manifold and subsequently
embedded as a configuration in Euclidean space using multidimensional
scaling.  A sparse support vector machine algorithm 
is then able to further reduce the dimension of the classifier
in Euclidean space by selecting only MDS dimensions
most useful for classification.

The results confirm the assertion that the geodesic
distance is inferior to the chordal metric for
applications of the Grassmann framework \cite{conway}.  
However, we have observed that the use of a smallest single angle
pseudo-metric improves classification accuracy even further.
The accuracies on the two class test problems grew to $100\%$ on 
the test data as we increased the dimension of the Grassmannian 
but only the pseudometric obtained 100\% classification on the challenging
soybean data.
The SSVM also provided additional insight into the pseudometric showing
that far fewer dimensions are required by the classifier.


\bibliographystyle{IEEEbib}
\bibliography{Bibliography}


\end{document}